\documentclass[11pt]{article}

\usepackage[preprint]{acl}

\usepackage{times}
\usepackage{latexsym}

\usepackage[T1]{fontenc}

\usepackage[utf8]{inputenc}

\usepackage{microtype}
\IfFileExists{inconsolata.sty}{\usepackage{inconsolata}}{} %
\usepackage{graphicx}
\usepackage{booktabs}
\usepackage{array}
\usepackage{multirow}
\usepackage{amsmath}
\usepackage{amssymb}
\usepackage{xcolor}
\usepackage{tikz}
\usetikzlibrary{arrows.meta,fit,shapes.geometric}
\usepackage{enumitem}
\usepackage{algorithm}
\usepackage{algpseudocode}

\title{Exploring Cross-Scenario Generality of Agentic Memory Systems: Diagnostics and a Strong Baseline}

\author{%
  \mdseries Zhikai Chen$^{1}$\thanks{\ Equal contribution.} \quad Jialiang Gu$^{2}$\footnotemark[1] \quad Junyu Yin$^{2}$\footnotemark[1] \quad Xianxuan Long$^{1}$ \quad Shenglai Zeng$^{1}$ \\
  \mdseries Xiaoze Liu$^{3}$ \quad Kai Guo$^{1}$ \quad Keren Zhou$^{2}$ \quad Jiliang Tang$^{1}$ \\
  \vspace{2pt} \\
  \mdseries $^{1}$Michigan State University \quad $^{2}$George Mason University \quad $^{3}$Purdue University \\
}

\begin{document}
\maketitle

\begin{abstract}
LLM agents accumulate histories that outgrow their context windows, motivating a growing literature on memory systems. Yet most existing designs are tuned to a single scenario (multi-session chat or a single trajectory format), and there is little evidence that they generalize across the heterogeneous trajectories agents encounter in deployment. We revisit eight memory systems plus an agentic harness for search problems, on five scenarios: single-turn QA, multi-session chat, agentic-trajectory QA, memory stress tests, and long-horizon agentic tasks. The harness, which self-manages flat text-file storage via tool calls, achieves the best cross-task ranking, suggesting that memory performance hinges on giving the agent active control over storage and retrieval rather than on a passive store behind a fixed pipeline. We instantiate this insight in \textbf{AutoMEM}, an agentic memory harness with a self-managed tool interface that achieves the best cross-scenario generality among the systems we evaluate.

\end{abstract}

\section{Introduction}

Large language models (LLMs) have evolved from single-turn chatbots~\citep{ouyang2022instructgpt} into general-purpose agents that plan, invoke tools, and execute long-horizon workflows~\citep{yao2023react, schick2024toolformer, wang2024survey}. Across these regimes, models accumulate historical information: knowledge corpora to ground responses~\citep{lewis2020rag}, codebases to navigate~\citep{jimenez2024swebench, cursor2025}, multi-step workflows to plan~\citep{xie2024travelplanner}, and user state to maintain~\citep{salemi2024lamp}. \textbf{Memory}, the mechanism that selects and organizes past information, is what lets an LLM act on more than the immediately available prompt.

Many memory systems have been proposed~\citep{hu2026memorysurvey}, yet whether the added complexity helps is increasingly contested~\citep{pollertlam2026beyondcontext,memoryarena2026}. The central problem is that \textbf{most existing memory systems are scenario-narrow}: each is tuned around a single regime, typically multi-session chat~\citep{longmemeval2026, maharana2024locomo} or one agentic-trajectory format, and the characteristics those regimes reward differ sharply~\citep{amabench2026}. Winning on one does not imply winning on others, and little evidence shows which designs generalize.

We therefore revisit existing designs through a \textbf{cross-scenario generality} lens: a practical memory system must work across the heterogeneous trajectories agents encounter in deployment, such as multi-session chat, code interactions, and browser logs. We also track token cost and latency to ensure generality is not bought at impractical cost~\citep{plugmem2026}.

We evaluate eight representative memory systems plus an \textbf{agent harness}~\citep{dciagent2025} designed for search across five task families: single-turn QA, multi-session QA, agentic-trajectory QA, memory stress tests, and long-horizon agentic tasks. These systems span long-context, note-based, multi-store hierarchical, graph-based, and RL-trained designs. The main observation is that even this off-the-shelf harness achieves the best cross-scenario generality: structured memories remain useful as tools, but on their own are too scenario-narrow to span real agentic trajectories. Spanning that diversity requires an \textbf{agentic memory harness} that self-manages memory through tool calls. We instantiate this in \textbf{AutoMEM}, which achieves the best cross-scenario generality across the systems we evaluate. Our key contributions:
\begin{enumerate}[leftmargin=*,topsep=1pt,itemsep=1pt,parsep=0pt,label=(\arabic*)]
\item A cross-scenario evaluation of eight memory systems plus an agent harness across five task families, with token cost and latency tracked alongside accuracy.
\item Empirical finding: existing memory systems struggle on agentic trajectories through two failure modes, a \textbf{representation-level} failure where build-time schemas drop step- and action-level evidence, and a \textbf{retrieval-level} failure where passive retrieval cannot surface evidence the storage retains; an agent harness that defers retrieval to query time achieves the best generality.
\item \textbf{AutoMEM}, an agent harness with a memory tool-calling interface, achieving $49.6\%$ improvement over the original agent harness on LoCoMo and $24.4\%$ in overall ranking.
\end{enumerate}

\section{Related Work}
\label{sec:preliminaries}

\subsection{Memory Design}
\label{sec:related-design}

Memory designs for LLM agents pursue two complementary goals: improving \textbf{quality} (accuracy, capability, generality) and improving \textbf{efficiency} (token cost, build cost, latency).

\paragraph{Quality.}
Memory design reduces to two coupled questions: how to store and update past information, and how to retrieve it. On the storage side, the central choice is granularity. The baseline stores raw chunked passages and retrieves them by similarity search~\citep{lewis2020rag, karpukhin2020dense}, on top of which three richer designs target different access patterns: atomic notes~\citep{mem0_2024, zhong2023memorybank, amem2025} make each fact an addressable, editable record, fitting evolving stable-fact workloads; OS-style tiered memory~\citep{packer2023memgpt, qian2025memorag, statelm2026} pages information through tool calls between bounded working and long-term tiers, fitting multi-session agents; and graph-based storage~\citep{edge2024graphrag, gutierrez2025hipporag2} makes entity-relation edges first-class, fitting multi-hop composition. 
On agentic tasks, however, single-granularity stores prove insufficient, and hierarchical or mixed-granularity stores~\citep{plugmem2026, memt2026, shu2026remem, amabench2026} are adopted at the cost of higher build overhead and token cost. 
Retrieval typically mirrors the storage form: graph traversal for graphs, semantic search for notes and passages, and tool-call paging for tiers. 
Beyond this default, two directions stand out: training an LLM tool-policy with RL to manage retrieval~\citep{memt2026, memrl2026, memoryr1, zhou2025mem1, wang2025memalpha, memagent2026} shifts memory work to post-training time at the cost of training compute, while multi-stage retrieval, the retrieval-side analog of mixed-granularity storage, ensembles multiple retrieval paths (e.g., graph traversal plus semantic search) and aggregates their candidates to recover what any single path misses, at the cost of more retrieval calls per query~\citep{plugmem2026, amabench2026}. 

\paragraph{Efficiency.}
Cost-focused work groups into three patterns. \textit{Compression} reduces content via gating, coreference resolution, and summarization~\citep{simplemem2026, chainofmemory2026, timem2026}. \textit{Offline delegation} shifts work to build time so retrieval becomes non-LLM~\citep{swiftmem2026, gutierrez2025hipporag2, plugmem2026}. \textit{Cost-tier routing} picks the cheapest module meeting a quality threshold~\citep{budgetmem2026}. All three are validated mainly on conversational QA, so transfer to agentic workloads is untested. We cover the first and leave the others as future work.

\subsection{Memory Benchmark}
\label{sec:related-benchmark}

Memory benchmarks fall into three categories. \textbf{QA benchmarks} test memory through QA over conversation history or long documents~\citep{maharana2024locomo, li2026locomoplus, longmemeval2026, lee2025realtalk, hu2025memoryagentbench, ai2025memorybench, pang2022quality, bai2025longbenchv2, hsieh2024ruler}. \textbf{Agentic QA benchmarks} ask questions about an agent's trajectory: AMA-Bench~\citep{amabench2026} uses machine-generated trajectories with causal dependencies; SAGE~\citep{sage2026} targets retrieval in deep research agents. \textbf{Real agentic task benchmarks} couple memory to actions so memory quality affects task completion; MemoryArena~\citep{memoryarena2026} runs memory-agent-environment loops. Existing benchmarks typically target a single category and rarely report token cost or latency, leaving no principled comparison along the cross-task, cost, and latency axes.

\section{Evaluation Design}
\label{sec:evaluation}

We measure each memory system on three axes: cross-scenario generality, token cost, and latency.

\paragraph{Tasks.} Table~\ref{tab:tasks} lists the five scenarios, which span the main deployment regimes for memory in agentic systems.

\begin{table}[t]
\centering
\footnotesize
\setlength{\tabcolsep}{4pt}
\renewcommand{\arraystretch}{0.95}
\begin{tabular}{@{}>{\raggedright\arraybackslash}p{1.7cm}>{\raggedright\arraybackslash}p{2.9cm}>{\raggedright\arraybackslash}p{2.6cm}@{}}
\toprule
\textbf{Scenario} & \textbf{Benchmark} & \textbf{Explanation} \\
\midrule
Personal chat assistant & LoCoMo~\citep{maharana2024locomo} & QA over personal daily affairs \\
Large-corpus retrieval & HotpotQA~\citep{yang2018hotpotqa} w/ Plugmem setup~\citep{plugmem2026} & QA over a corpus beyond context size \\
Trajectory recall & AMABench-\{ALF, Web, T2SQL\}~\citep{amabench2026} & QA over agent trajectories \\
Memory stress tests & MemoryAgentBench-\{AR, TTL, LRU, CR\}~\citep{hu2025memoryagentbench} & Accurate retrieval, test-time learning, long-range understanding, conflict resolution \\
Real agentic tasks & ALFWorld~\citep{shridhar2021alfworld}, MemoryArena-shop / -travel~\citep{memoryarena2026} & Real-world agentic tasks \\
\bottomrule
\end{tabular}
\caption{The five task scenarios.}
\label{tab:tasks}
\end{table}

\paragraph{Memory baselines.} We evaluate eight memory systems plus an agent harness baseline~\citep{dciagent2025} originally designed for search (Table~\ref{tab:baselines}), covering the architectural design space of \S\ref{sec:related-design}. For agentic tasks we use a ReAct agent~\citep{yao2023react} with task-tuned prompts.

\begin{table}[t]
\centering
\footnotesize
\setlength{\tabcolsep}{4pt}
\renewcommand{\arraystretch}{0.95}
\begin{tabular}{@{}>{\raggedright\arraybackslash}p{1.6cm}>{\raggedright\arraybackslash}p{3.4cm}>{\raggedright\arraybackslash}p{2.2cm}@{}}
\toprule
\textbf{Category} & \textbf{Method(s)} & \textbf{Storage primitive} \\
\midrule
No memory & Long context & Full prompt \\
Note-based & SimpleMem~\citep{simplemem2026} & Atomic notes \\
Multi-store & PlugMem~\citep{plugmem2026}, LightMem~\citep{lightmem2025} & Episodic $+$ semantic $+$ procedural stores \\
Graph-based & HippoRAG~\citep{gutierrez2025hipporag2}, AMA-Agent~\citep{amabench2026} & Entity-fact KG / turn-indexed graph \\
Agentic harness & DCI-Lite~\citep{dciagent2025}, DCI-Lite$+$Sum~\citep{dciagent2025} & Raw trajectory $+$ grep/read agent \\
RL-trained & Mem-T~\citep{memt2026}, MemRL~\citep{memrl2026} & Trained retrieval / write policy \\
\bottomrule
\end{tabular}
\caption{Memory baselines grouped by architectural category.}
\label{tab:baselines}
\end{table}

\paragraph{Metrics.} For LoCoMo, HotpotQA, AMABench, and MemoryAgentBench we report a Qwen3-32B LLM-judge score~\citep{amabench2026}, since token F1 mis-scores long-form answers on verbose benchmarks (Appendix~\ref{app:locomo-f1-example}). For agentic tasks we report environment outcomes: success rate on ALFWorld; process score on MA-Shop (fraction of subtasks solved); and c-sPS on MA-Travel, a corrected MemoryArena sPS~\citep{memoryarena2026} that only scores slots the user asks about in the current session.

\paragraph{Backbones.} The default backbone is Qwen3-32B~\citep{yang2025qwen3} for both construction and answering, with Qwen2.5-7B-Instruct on ALFWorld (Qwen3-32B solves most episodes trivially). The embedding model is Qwen3-Embedding-4B throughout. A Qwen3-4B-Instruct ablation is in Appendix~\ref{app:qwen3-4b-protocol}.

\section{Experimental Results}
\label{sec:results}

\subsection{Overview}
\label{sec:results-overview}

\begin{figure*}[t]
\centering
\includegraphics[width=\textwidth]{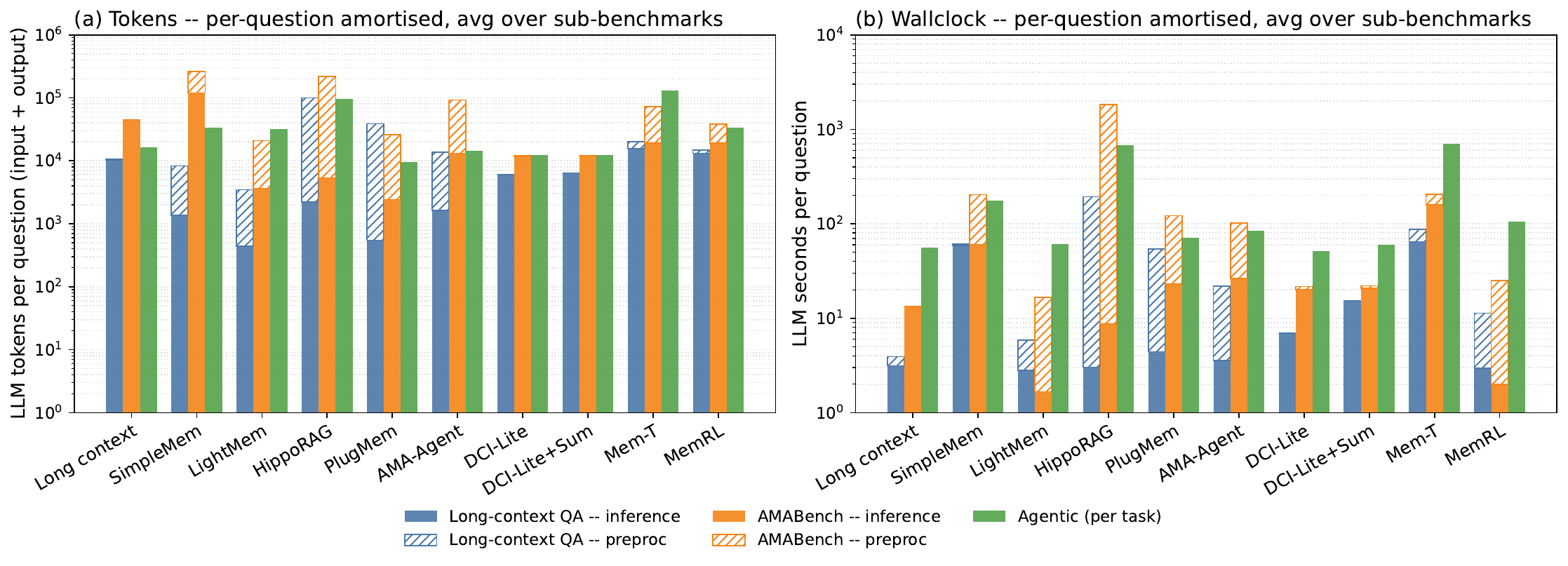}
\caption{\textbf{Per-method preprocessing vs.\ inference cost per question}, averaged across sub-benchmarks within each task category.}
\label{fig:efficiency}
\end{figure*}

We report performance across benchmarks in Table~\ref{tab:generality} and per-method preprocessing vs.\ inference cost in Fig.~\ref{fig:efficiency}. The following high-level patterns emerge:

\begin{enumerate}[leftmargin=*,topsep=4pt,itemsep=3pt,parsep=0pt,label=(\arabic*)]
\item \textbf{No method dominates.} Every index-based method (one that pre-builds a structured store such as a graph, summary notes, or multi-store cache) lags long context on at least one benchmark; DCI-Lite holds the best generality rank. The shortfall of index-based methods is most visible on agentic-trajectory QA (\S\ref{sec:results-agentic-qa}).
\item \textbf{Long context is stronger than commonly assumed}~\citep{hu2025memoryagentbench,amabench2026} and remains cost-competitive: heavy-index methods such as HippoRAG cannot recover preprocessing cost unless many future queries hit the same store, while lighter alternatives such as LightMem keep the build cheap but score lower than long context. Neither Pareto-dominates the baseline.
\item \textbf{On dynamic agentic tasks, most methods converge} within sampling variance.
\item \textbf{Token efficiency $\ne$ system efficiency.} HippoRAG and AMA-Agent issue tokens as long chains of small serial LLM calls with KV-cache misses, inflating wall-clock and GPU contention far beyond what raw token counts suggest. This reflects a memory system's infrastructure affinity, which token counts alone do not capture; per-method numbers are in Appendix~\ref{par:sys-eff}.
\end{enumerate}

The rest of \S\ref{sec:results} follows one thread: when to commit memory structure at build time versus query time. \S\ref{sec:results-agentic-qa} diagnoses why early commitment fails on agentic QA; \S\ref{sec:results-harnessing} shows the agent harness wins by deferring that commitment to query time; \S\ref{sec:results-indexing} characterizes when the build-time cost still pays off; \S\ref{sec:results-agentic-gap} closes with the scope limit: on dynamic agentic tasks, no memory design closes the gap.

\begin{table*}[t]
\centering
\small
\resizebox{\textwidth}{!}{%
\begin{tabular}{lccccccccccccc}
\toprule
\textbf{Method} & \textbf{LoCoMo} & \textbf{A-ALF} & \textbf{A-Web} & \textbf{A-SQL} & \textbf{HQA} & \textbf{AR} & \textbf{TTL} & \textbf{LRU} & \textbf{CR} & \textbf{ALFW} & \textbf{MA-S} & \textbf{MA-T} & \textbf{Gen.} \\
 & \textit{LLM judge} & \textit{LLM judge} & \textit{LLM judge} & \textit{LLM judge} & \textit{LLM judge} & \textit{LLM judge} & \textit{LLM judge} & \textit{LLM judge} & \textit{LLM judge} & \textit{SR} & \textit{PS} & \textit{c-sPS} & \textbf{rank} \\
\midrule
Long context    & 61.5          & 26.9          & \textbf{43.9} & \textbf{35.4} & 58.0          & \textbf{82.0} & 52.0          & 6.0           & 12.0          & \textbf{46.7} & 32.8          & 12.8          & 4.00 \\
SimpleMem       & 56.5          & 8.3           & 18.2          & 13.8          & 70.5          & 56.0          & 46.0          & \textbf{26.0} & 20.0          & 43.3          & 26.7          & 17.7          & 6.83 \\
LightMem        & 62.0          & 9.4           & 25.3          & 15.6          & 78.5          & 62.0          & 60.0          & 10.0          & 16.0          & 43.3          & \textbf{38.3} & 15.5          & 5.11 \\
HippoRAG        & 58.0          & 10.9          & 28.6          & 23.0          & 83.5          & 74.0          & 58.0          & 10.0          & 14.0          & 40.0          & 31.7          & \textbf{18.8} & 4.17 \\
PlugMem         & 53.0          & 25.0          & 16.7          & 18.1          & 63.5          & 64.0          & 58.0          & 10.0          & \textbf{24.0} & 33.3          & 28.3          & 17.7          & 6.94 \\
AMA-Agent       & 54.5          & 10.0          & 31.0          & 22.9          & 78.0          & 66.0          & 42.0          & 6.0           & 12.0          & 31.7          & 30.0          & 18.0          & 6.61 \\
DCI-Lite        & 45.0          & \textbf{32.2} & 37.2          & 25.5          & 87.5          & 60.0          & 54.0          & 10.0          & 14.0          & 41.7          & 33.3          & \textbf{18.8} & \textbf{3.67} \\
DCI-Lite+Sum    & 46.5          & 30.8          & 38.1          & 27.2          & \textbf{89.0} & 58.0          & 48.0          & 6.0           & 16.0          & 40.0          & 30.6          & 18.4          & 4.28 \\
Mem-T           & \textbf{66.5} & 26.0          & 32.1          & 10.0          & 63.5          & 38.0          & \textbf{66.0} & 2.0           & 10.0          & 38.3          & 30.0          & 18.4          & 6.06 \\
MemRL           & 47.5          & 5.4           & 29.9          & 18.8          & 62.0          & 64.0          & 54.0          & 4.0           & 6.0           & 36.7          & 33.3          & 14.4          & 7.33 \\
\midrule
Variance        & 1.4 & 1.7 & 0.2 & 0.6 & 1.0 & 2.8 & 3.0 & 1.8 & 2.0 & 3.9 & 4.2 & 1.1 & -- \\
\bottomrule
\end{tabular}
}%
\caption{Main results: per-method accuracy across all benchmarks; best per column in \textbf{bold}. A-ALF/A-Web/A-SQL are AMABench domains; HQA is HotpotQA; AR/TTL/LRU/CR are MemoryAgentBench~\citep{hu2025memoryagentbench} subtasks; ALFW is ALFWorld; MA-S/MA-T are MemoryArena shopping/travel. The Gen.\ column is the mean fractional rank across benchmarks, with MemoryAgentBench counted once by averaging AR/TTL/LRU/CR. The Variance row reports the sampling-noise floor in percentage points.}
\label{tab:generality}
\end{table*}

\subsection{Where index-based memory fails on agentic QA}
\label{sec:results-agentic-qa}

\begin{figure}[t]
\centering
\begin{tikzpicture}[
  >=Stealth,
  trajstep/.style={draw, rounded corners=2pt, fill=gray!10,
    minimum width=1.25cm, minimum height=0.5cm,
    align=center, font=\scriptsize},
  kgnode/.style={draw, ellipse, fill=blue!10, font=\scriptsize, inner sep=2pt},
  envnode/.style={draw, circle, fill=teal!10, font=\scriptsize,
    minimum size=0.55cm, inner sep=0pt},
]
\node[font=\scriptsize\itshape] at (3.0,0.55) {Raw trajectory};
\node[trajstep] (s5) at (1.5,0) {\texttt{scroll}\\Step 5};
\node[trajstep] (s6) at (3.0,0) {\texttt{scroll}\\Step 6};
\node[trajstep] (s7) at (4.5,0) {\texttt{go\_back}\\Step 7};
\draw[->] (s5) -- (s6);
\draw[->] (s6) -- (s7);
\node[font=\scriptsize\bfseries, anchor=east] at (0.95,-1.15) {HippoRAG:};
\node[kgnode] (hc) at (2.0,-1.15) {black case};
\node[kgnode] (hp) at (3.7,-1.15) {\$11.99};
\draw[->] (hc) -- node[above,font=\tiny]{has\_price} (hp);
\node[font=\scriptsize, color=red!70!black, anchor=west] at (4.55,-1.15) {step/action: lost};
\node[font=\scriptsize\bfseries, anchor=east] at (0.95,-2.05) {AMA-Agent:};
\node[envnode] (e5) at (1.5,-2.05) {e@5};
\node[envnode] (e6) at (2.5,-2.05) {e@6};
\node[envnode] (e7) at (3.5,-2.05) {e@7};
\draw[->] (e5) -- node[above,font=\tiny]{scroll} (e6);
\draw[->] (e6) -- node[above,font=\tiny]{scroll} (e7);
\draw[->] (e7) -- node[above,font=\tiny]{go\_back} ++(0.7,0);
\node[font=\scriptsize, color=green!50!black, anchor=west] at (4.55,-2.05) {step/action: kept};
\end{tikzpicture}
\caption{Schema loss on an AMABench-Web question (\textit{``which step performed the return from the product page, and how many scrolls happened before it?''}). HippoRAG keeps entity-relation facts but discards step indices and actions; AMA-Agent's turn-indexed graph keeps both.}
\label{fig:schema-loss}
\end{figure}
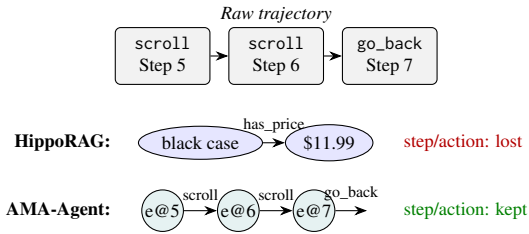

Indexing-based methods perform poorly on agentic QA (Table~\ref{tab:generality}, AMABench columns), with several scoring less than half the long-context baseline despite building richer structures. To localize the failure, we ablate the pipeline and ask whether the answer is still recoverable before the normal retrieval interface is applied. HippoRAG2 extracts an entity-fact KG; AMA-Agent attaches per-turn state reports to a turn-indexed causality graph with action-labeled edges; PlugMem keeps three parallel stores (episodic raw trace, semantic entity facts, procedural subgoal summaries) behind a mode-routing retriever. We sample 60 questions across the three AMABench domains and use Codex with gpt-5.5 (extra high reasoning) to check this oracle storage answerability. Methods designed for personal-chat or Wikipedia-style assistants hit a hard representational ceiling (Table~\ref{tab:answerability-ceiling}); AMA-Agent and PlugMem show a different pattern: the answer is often still present somewhere in storage, but the deployed retrieval interface fails to surface it reliably.

\begin{table}[t]
\centering
\footnotesize
\begin{tabular}{lccccc}
\toprule
\textbf{Method} & \textbf{A} & \textbf{B} & \textbf{C} & \textbf{D} & \textbf{Overall} \\
\midrule
HippoRAG  &  19 &   7 &   \phantom{0}0 &  14 &  10 \\
SimpleMem &  38 &  40 &  33 &  29 &  35 \\
AMA-Agent & 100 &  80 &  67 &  86 &  83 \\
PlugMem   & 100 & 100 &  93 & 100 &  98 \\
\bottomrule
\end{tabular}
\caption{Oracle storage answerability (\%). The probe asks whether the answer is still recoverable from the method's stored representation, or from the raw episodic trace when the method keeps one; it is therefore an upper bound on what the retrieval interface can deliver, not an end-to-end retrieval score. Question types are AMABench's: A state-value lookup, B causal inference, C causal-chain, D constraint comparison.}
\label{tab:answerability-ceiling}
\end{table}

A deeper investigation reveals two failure modes. The first is a \textbf{representation-level} failure (Figure~\ref{fig:schema-loss}): HippoRAG's step indices and action labels never enter the entity-relation KG because they are not entity-relation facts, so the data is gone before retrieval and no answerer can recover it; the only fix is to widen the schema at build time.

The second is a \textbf{retrieval-level} failure: storage isn't where the answer is lost. Table~\ref{tab:answerability-ceiling} shows that PlugMem keeps over $98\%$ of the answer information (its episodic store holds the dense raw trace), so an oracle reading the whole memory could answer almost any question. What fails is \textbf{passive retrieval}: a router picks one branch by question-embedding similarity and returns a compact note; when the answer lives in step-level detail, the note never carries, the raw trace stays in storage, and the answerer has no second chance to ask for it. An active agent-driven retrieval interface would close this gap: when the returned evidence is not enough, the agent can issue a follow-up call and fetch what is missing.

A further issue is \textit{inflexibility}: index-based memory commits to a schema at build time, so the designer must anticipate which relation types future questions will probe. To accommodate agentic trajectories, AMA-Agent and PlugMem must add more complexity to the graph and include more node and relation types, which is unlikely to scale as agentic scenarios grow more diverse. Letting an agent actively write the query partially addresses this; the intrinsic problems of index-based memory for agentic use cases are left for future study.

\subsection{When the agent harness wins}
\label{sec:results-harnessing}

Both failures in \S\ref{sec:results-agentic-qa} stem from committing memory structure at build time. We now turn to the design that defers commitment to query time: DCI-Lite, an agent harness that keeps evidence in raw text and lets the LLM compose a per-question retrieval plan via \texttt{grep} and \texttt{read}, holds the best generality rank in our suite. Compared to index-based methods that commit a schema at build time and pay for that choice with the inflexibility of \S\ref{sec:results-agentic-qa}, the harness inverts the binding order: the LLM sees the actual question first, then decides at runtime which slice of raw evidence to read and how to aggregate it, shifting the schema-commitment cost from a one-time frozen choice at build time to a per-question choice at query time informed by the question itself. Runtime overhead is comparable to LongCtx since both route the retrieved evidence through an answer LLM; what DCI-Lite buys with its grep--read loop is generality across regimes, not raw token savings on any single bench. The cost win is clearer against heavy-index methods: HippoRAG pays a ${\sim}130$K-per-question amortized build on Long-context QA versus DCI-Lite's near-zero build (Fig.~\ref{fig:efficiency}).

Table~\ref{tab:dci-probes} reports three signals behind DCI's lead, with one probe per signal while pinning the other two. \textbf{Predicate complexity:} HippoRAG indexes Wikipedia entities by name, and each HotpotQA question hinges on two such entities. When the question names both entities, the index lookup is trivial; when it refers to them indirectly, the index must first recover the entity names before it can do anything. Structured indexing tends to win in the first case, DCI in the second. \textbf{Corpus--window regime:} the harness becomes valuable when a single read of the corpus is impractical. \textbf{Schema expressiveness:} the index has a hard representational ceiling that no amount of retrieval quality can lift.

\begin{table}[t]
\centering
\small
\begin{tabular}{lr}
\toprule
\textbf{Probe (DCI vs.\ baseline)} & \textbf{$\Delta$\,pp} \\
\midrule
\multicolumn{2}{l}{\textit{Predicate complexity} (vs.\ HippoRAG, HQA)} \\
\quad neither key entity directly in question & $+11.3$ \\
\quad both key entities directly in question  & $-3.1$  \\
\midrule
\multicolumn{2}{l}{\textit{Corpus--window regime} (vs.\ LongCtx)} \\
\quad LoCoMo, ${\sim}26$k tokens (fits window)        & $-16.5$ \\
\quad HotpotQA, $9{,}811$ passages (exceeds window)   & $+29.5$ \\
\midrule
\multicolumn{2}{l}{\textit{Schema expressiveness} (vs.\ HippoRAG)} \\
\quad AMABench-ALF, 10\% encodable    & $+21.3$ \\
\bottomrule
\end{tabular}
\caption{Three controlled probes for DCI's generality lead. $\Delta$ is DCI minus the named baseline.}
\label{tab:dci-probes}
\end{table}

\subsection{When indexing pays off}
\label{sec:results-indexing}

We now study when an index-based method is still worth its build cost, along two axes. From the \textbf{efficiency} side, we ask when the per-question token cost of an index drops below the no-build baselines once enough queries amortize the build. From the \textbf{performance} side, we ask which task properties make index-based methods a good fit in the first place. The combined picture is that an index pays off only when both axes are satisfied: the workload amortizes the build, \textit{and} the task structure rewards build-time aggregation.

\begin{figure}[t]
\centering
\includegraphics[width=\columnwidth]{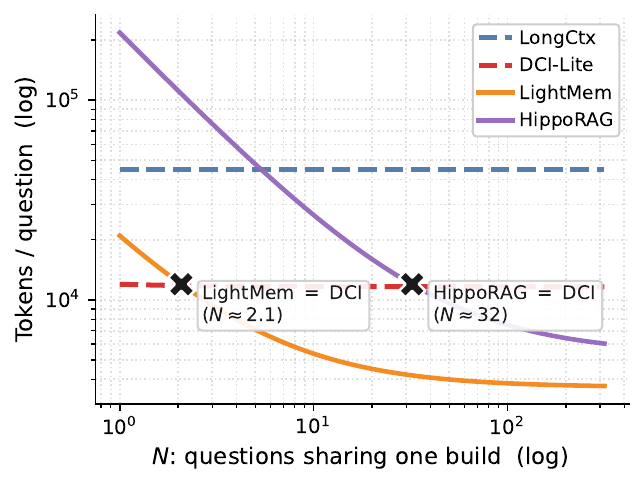}
\caption{Per-question token cost vs.\ amortisation $N$ on AMABench. LightMem crosses DCI-Lite at $N{\approx}2$; HippoRAG at $N{\approx}32$.}
\label{fig:pareto}
\end{figure}

\paragraph{Efficiency axis: the amortization effect.} An index-based method pays its build cost once and amortizes it across all questions that hit the same store. As the number of shared questions $N$ grows, its per-question cost drops; no-build methods (LongCtx, DCI-Lite) stay flat. Fig.~\ref{fig:pareto} makes this concrete on AMABench: LightMem crosses DCI-Lite at $N{\approx}2$, and even HippoRAG, despite its much larger build, catches up at $N{\approx}32$.

\paragraph{Performance axis: when does indexing help?} Being cheap is not the same as being useful: Table~\ref{tab:generality} shows index-based methods are dominated in accuracy on most benchmarks, even where the build cost is amortized away by enough queries. We first characterize what an index does at storage time, then identify the workload property that makes that work valuable. The index-based methods in our suite come in two flavors. \textit{Graph-based} methods like HippoRAG2 encode \textit{explicit} relations between stored units (entity-fact triples, turn-indexed causal edges, or multi-store mode tags) at build time. \textit{Semantic note-based} methods like SimpleMem compress multiple turns into LLM-summarized notes and embed the notes, encoding cross-unit content \textit{implicitly} in the summary text. The two flavors look different at the storage layer, but they share one defining property: both pay a build-time cost to \textbf{aggregate information across raw units}, so a downstream query can answer with a single retrieval rather than reconstruct the relation on the fly. This shared property, not the specific mechanism, controls when indexing pays off.

Aggregation only pays off when the workload demands it. We measure that demand by the \textbf{structural rate}: the fraction of questions whose answer cannot be recovered from any single retrieval unit (passage, dialog turn, or trajectory step) under flat retrieval (grep, BM25, or dense top-$5$), and therefore requires combining content across units. It upper-bounds the room build-time aggregation that has to help.

\begin{table}[h]
\centering
\footnotesize
\renewcommand{\arraystretch}{1.15}
\resizebox{\columnwidth}{!}{%
\begin{tabular}{@{}lccc@{}}
\toprule
                  & \textbf{Struct.} & \multicolumn{2}{c}{\textbf{Avg.\ $\Delta$ vs.\ DCI-Lite (pp)}} \\
\cmidrule(lr){3-4}
\textbf{Benchmark} & \textbf{rate}    & \textbf{graph} & \textbf{semantic} \\
\midrule
\textbf{LoCoMo}            & $\mathbf{60.7\%}$ & $\mathbf{+13.0}$ & $\mathbf{+14.2}$ \\
\addlinespace[2pt]
HotpotQA                   & $17.7\%$          & $-15.4$          & $-13.0$ \\
AMABench-ALF               & $10.3\%$          & $-14.2$          & $-23.4$ \\
AMABench-T2SQL             & $\phantom{0}8.6\%$ & $\phantom{0}-7.0$ & $-10.8$ \\
\bottomrule
\end{tabular}%
}%
\caption{Structural rate predicts the head-to-head between indexed methods and the no-index harness DCI-Lite. Accuracy $\Delta$ in pp, averaged over methods in each flavor (graph: HippoRAG, PlugMem, AMA-Agent, Mem-T; semantic: LightMem, SimpleMem).}
\label{tab:recoverability}
\end{table}

\paragraph{Observation and implication.} Two corpus-side properties jointly gate when indexing pays off, and they play distinct roles. \textit{Schema expressiveness} (\S\ref{sec:results-harnessing}) gates whether indexing \textit{can} help at all: if the schema cannot encode the predicates the questions probe, no amount of structural demand translates into accuracy gains (the AMABench-ALF failure mode). \textit{Structural rate} (this section) gates whether indexing is \textit{needed}: even with a fitting schema, indexing only adds value when many questions require cross-unit aggregation a flat retriever cannot do. Both must hold for indexed methods to beat the no-index harness.

\subsection{Scope: dynamic agentic tasks are policy-limited}
\label{sec:results-agentic-gap}

The preceding subsections assume static-style QA, where the right memory design can improve accuracy. On dynamic agentic tasks this assumption breaks: no memory choice in our suite closes the gap, regardless of when it commits. On ALFWorld, no memory method consistently improves the long-context baseline, while parameter-level post-training (GRPO~\citep{shao2024deepseekmath} on Qwen2.5-7B) does. To localize where memory is blocked, we hand-craft a \textbf{golden procedural memory} by distilling the rules that win ALFWorld games from a small set of expert trajectories into a single static rule sheet injected via the system prompt, and compare four conditions (Table~\ref{tab:agentic-gap}). The oracle closes only part of the gap to the trained actor.

\begin{table}[t]
\centering
\small
\begin{tabular}{lr}
\toprule
Condition (Qwen2.5-7B actor) & ALFWorld SR \\
\midrule
Vanilla long-context (no memory)   & 46.7\% \\
Best memory$^{\dagger}$    & 43.3\% \\
Golden procedural memory             & 71.7\% \\
GRPO post-trained                  & 86.7\% \\
\bottomrule
\end{tabular}
\caption{ALFWorld success rate with Qwen2.5-7B as the actor backbone. $^{\dagger}$Strongest of all memory methods on the ALFWorld column of Table~\ref{tab:generality} (SimpleMem / LightMem at 43.3\%).}
\label{tab:agentic-gap}
\end{table}

The gap splits into two pieces: one that the golden memory can close, and one that it cannot.

\textbf{Gap~1: over-entropic action posterior.} Without a rule in context, the actor wanders across too many candidate actions and runs out of the 50-step budget (all 32 failed trajectories hit the cap). Golden memory fixes this by spelling out the recipe for each task family, and the lift concentrates on tasks whose solution can be written down as a short procedure. Memory systems fail to deliver the same lift because reflections written from past trajectories capture episode-specific events rather than reusable procedures. An expert is needed to distill these memories from trajectories.

\textbf{Gap~2: state-dependent decisions text rules cannot encode.} The remaining 17 of 32 failures persist even with the golden recipe in context. Inspecting them, the correct next action depends on the current environment state (e.g., which receptacles are currently open, what the agent is already holding) rather than on a fixed recipe step. A text rule cannot enumerate the right branch for every possible state, so the actor still picks the wrong action at these decision points. The rule can also actively mislead: when its vocabulary surface-matches the task description, the actor mechanically follows it and triggers unnecessary action sequences. Closing this gap requires reshaping the actor's behavior at exactly those state-conditional decision points.

These two gaps connect to two open lines on memory-based agent self-improvement. \textbf{Gap~1} explains why an agent struggles to self-summarize its own procedural memory: a weaker actor cannot reliably distill canonical recipes from its own trajectories, so the propositional content must be sourced from a higher-capability LLM oracle and distilled into the actor (as in SkillRL~\citep{xia2026skillrl}) before it can produce the lift. \textbf{Gap~2} delineates the capability ceiling of self-evolving agents whose only update channel is non-parametric memory: such agents cannot move past behavior already implicit in the actor's prior. Going beyond this ceiling requires parametric updates such as the RL-based self-evolution in OpenClaw-RL~\citep{openclaw_rl}. Mem-T and MemRL sit on the non-parametric side: they post-train a memory controller on top of an unchanged actor, so they remain bounded by Gap~2.

\section{AutoMEM: A Design Inspired by Our Findings}
\label{sec:automem}

\providecommand{\efftodo}[1]{\textcolor{red}{\textbf{[#1]}}}

The experiments in \S\ref{sec:results} suggest that the agent harness performs well on most benchmarks and only loses ground when an answer must be pieced together from many records, where an index can close the gap if its schema matches the questions. The natural design is to migrate the indexed memory component into the harness.

We instantiate this as an agentic loop (Algorithm~\ref{alg:automem}). At each step, the agent (1) picks one action from a four-item menu: \texttt{rg} for regex over raw records, \texttt{read} for fetching a record by identifier, \texttt{dump} for sending a short corpus to the answer LLM, and \texttt{index\_query} for a typed query against a pre-built index. (2) A planner LLM writes the call, an executor runs it, and the evidence is appended to a running trace. (3) An LLM judge inspects the trace; on sufficiency, an answerer composes the final answer, on insufficiency, the judge returns a hint naming what is missing, and the planner is re-invoked. One LLM backbone serves all roles. By default, \texttt{index\_query} runs against a Cypher graph~\citep{francis2018cypher} built from LightMem's write-time summary index with edges \texttt{SAID\_BY}, \texttt{OF\_TOPIC}, \texttt{FOLLOWS}; a PlugMem-graph variant uses PlugMem's tag/semantic/episodic stores with edges \texttt{HAS\_TAG}, \texttt{MENTIONS}, \texttt{SHARES\_TAG}.
\begin{algorithm}[t]
\small
\caption{AutoMEM: agentic memory-access loop}\label{alg:automem}
\begin{algorithmic}[1]
\Require memory store $M$,\; question $q$,\; step budget $K_{\max}$
\Require action set $\mathcal{A}=\{\texttt{rg},\,\texttt{read},\,\texttt{dump},\,\texttt{index\_query}\}$
\Statex \textit{\footnotesize\textcolor{gray}{// $\mathit{trace}$: evidence so far. $\mathit{hint}$: judge feedback.}}
\State $\mathit{trace} \gets [\,]$;\; $\mathit{hint} \gets \varnothing$
\For{$k = 1, \dots, K_{\max}$}
  \Statex \quad\textit{\footnotesize\textcolor{gray}{// Plan: 1 LLM call, emits one action $\in\mathcal{A}$.}}
  \State $a_k \gets \textsc{Plan}(q,\, \mathit{trace},\, \mathit{hint})$
  \Statex \quad\textit{\footnotesize\textcolor{gray}{// Execute: tool call on $M$, no LLM.}}
  \State $o_k \gets \textsc{Execute}(M,\, a_k)$
  \State $\mathit{trace} \gets \mathit{trace}\;\Vert\;(a_k, o_k)$
  \Statex \quad\textit{\footnotesize\textcolor{gray}{// Judge: 1 LLM call. Skipped when $K_{\max}{=}1$ (single-pass).}}
  \State $v \gets \textsc{Judge}(q,\, \mathit{trace})$
  \If{$v.\mathit{sufficient}$} \textbf{break}
  \Else\; $\mathit{hint} \gets v.\mathit{reason}$
  \EndIf
\EndFor
\Statex \textit{\footnotesize\textcolor{gray}{// Answer: 1 LLM call composes the final answer.}}
\State \Return $\textsc{Answer}(q,\, \mathit{trace})$
\end{algorithmic}
\end{algorithm}

\paragraph{Results.}
AutoMEM beats DCI-Lite and long context on LoCoMo (Table~\ref{tab:automem}); the LightMem-graph default leads the PlugMem-graph variant by $1.3$ pp. On LoCoMo, AutoMEM ($67.3$) improves over DCI-Lite ($45.0$) by $+22.3$ pp, over long context ($61.5$) by $+5.8$ pp, and edges the previously strongest method Mem-T ($66.5$) by $+0.8$ pp. AutoMEM also tops the suite on average rank ($3.10$, vs.\ $4.10$ for DCI-Lite once inserted into Table~\ref{tab:generality}); the largest single-column lift sits on LoCoMo, the benchmark with the highest structural rate.

\begin{table}[h]
\centering
\small
\begin{tabular}{lcc}
\toprule
\textbf{Method} & \textbf{LoCoMo} & \textbf{Avg.\ rank} $\downarrow$ \\
\midrule
\textbf{AutoMEM}                  & \textbf{67.3} & \textbf{3.10} \\
AutoMEM (PlugMem graph)          & 66.0          & 3.21 \\
AutoMEM (single-pass)            & 63.5          & 3.38 \\
Long context                     & 61.5          & 4.93 \\
DCI-Lite                         & 45.0          & 4.10 \\
\bottomrule
\end{tabular}
\caption{LoCoMo accuracy and \textbf{average fractional rank across all twelve task columns of Table~\ref{tab:generality}}. LoCoMo is one of the benchmarks where DCI-Lite itself underperforms (45.0, below the long-context baseline at 61.5). Each AutoMEM row's rank is computed by inserting that variant as the new method into Table~\ref{tab:generality}.}
\label{tab:automem}
\end{table}

\paragraph{General algorithm.}
Algorithm~\ref{alg:automem} defines the canonical AutoMEM design: a multi-step plan--execute--judge loop that can re-query memory when the judge finds the trace insufficient. This iterative form is what delivers the accuracy in Table~\ref{tab:automem}. On top of long context's single LLM call, the loop issues three to four calls per question (planner, optional dump, judge, answerer), and the outer loop can issue further calls when the judge requests more evidence (Table~\ref{tab:automem-efficiency}).

\paragraph{Further efficiency improvements.}
On top of Algorithm~\ref{alg:automem}, we layer two latency-oriented optimizations that retain most of its accuracy at a fraction of the token cost, and apply them identically across the three QA benchmarks.
(i)~\textit{KV-cache-friendly chat layout} \textemdash{} amortise prefill across the multi-call loop. The stable instruction block lives in the system message; the user message places the corpus catalog and graph-schema hint before the per-question delta. On Qwen3-32B / vLLM, this lifts the in-run prefix-cache hit rate from $5.4\%$ to over $50\%$, so most multi-call overhead resolves to cache reads.
(ii)~\textit{Architectural single-pass cap} \textemdash{} replace the loop with one parallel pass. As a deployment ablation of Algorithm~\ref{alg:automem}, we fix $K_{\max}{=}1$, disable the \texttt{dump} branch, and skip the judge--replan step; the planner issues a single call, runs $\texttt{rg}$, $\texttt{read}$, and $\texttt{index\_query}$ in parallel, and returns. At a small accuracy cost ($-3.8$ pp on LoCoMo), single-pass cuts query-time tokens by $66$--$72\%$ across the three QA benchmarks (full per-benchmark numbers in Table~\ref{tab:automem-efficiency}). At that price, it is usually cheaper than long context and within $\pm 1.6\times$ of DCI-Lite's single-call cost.

\begin{table}[h]
\centering
\footnotesize
\setlength{\tabcolsep}{4pt}
\renewcommand{\arraystretch}{1.05}
\resizebox{\columnwidth}{!}{%
\begin{tabular}{l r r r r r r}
\toprule
& \multicolumn{2}{c}{\textbf{LoCoMo}} & \multicolumn{2}{c}{\textbf{HotpotQA}} & \multicolumn{2}{c}{\textbf{AmaBench}} \\
\cmidrule(lr){2-3}\cmidrule(lr){4-5}\cmidrule(lr){6-7}
\textbf{Method} & build & query & build & query & build & query \\
\midrule
Long context             & 0K    & 19.8K & 0K     & 0.9K & 0K    & 32.8K \\
DCI-Lite                 & 0K    & 10.8K & 0K     & 1.3K & 0K    & 8.8K  \\
HippoRAG~v2              & 13.5K & 3.5K  & 181.1K & 0.9K & 163.1K & 12.8K \\
AutoMEM (full loop)      & 5.5K$^{\dagger}$ & 24.8K & 0.6K$^{\dagger}$ & 5.9K & 20.8K$^{\dagger}$ & 25.3K \\
AutoMEM (single-pass)    & 5.5K$^{\dagger}$ & 6.9K & 0.6K$^{\dagger}$ & 2.0K & 20.8K$^{\dagger}$ & 8.2K \\
\bottomrule
\end{tabular}}%
\caption{Per-question LLM token cost; preprocessing is amortised the same way as Fig.~\ref{fig:efficiency}. $^{\dagger}$AutoMEM's build equals LightMem-graph's preprocessing; the PlugMem-graph variant substitutes PlugMem's ($24.2$K, $52.2$K, $41.5$K).}
\label{tab:automem-efficiency}
\end{table}

\section{Conclusion}
\label{sec:conclusion}

We revisit memory systems for LLM agents through a cross-scenario generality and cost lens and draw: (1) no existing memory system wins across all five task families, while an off-the-shelf agent harness achieves the best generality; (2) index-based methods fail on agentic-trajectory QA in two ways: build-time schemas drop step- and action-level evidence (storage), and passive retrieval cannot surface evidence the storage retains (retrieval); (3) reaching the cost--accuracy frontier requires combining the agent harness with selective indexing, which we instantiate in AutoMEM.

We see two potential future directions: (i)~\textit{a memory DSL that automates schema design} \textemdash{} a DSL over memory primitives could replace verbose Cypher and let the schema be meta-learned per workload~\citep{xiong2026metamem}; (ii)~\textit{dedicated infrastructure for memory-bound agents} \textemdash{} serial prefill across plan/judge/answer dominates wall-clock (\S\ref{par:sys-eff}), and a program-aware serving stack that batches these calls and reuses prefix caches~\citep{kang2026thunderagent} could close most of the latency gap to long-context QA.

\section*{Limitations}

\paragraph{No tiered budget routing.}
AutoMEM runs plan, judge and answer composition on the same backbone, but the three steps differ in difficulty: planning over a small corpus catalog and judging a candidate answer's sufficiency are short, low-entropy decisions, while answer composition is the step that needs the strong model's reasoning. A per-step router that sends plan and judge to a cheap tier and reserves the strong backbone for answer composition would, by the call-count breakdown in our efficiency table, cut the AutoMEM per-question LLM cost by roughly $2$ to $3$ times. We do not implement this router and report uniform-backbone numbers throughout, so the reported cost of AutoMEM should be read as an upper bound.

\paragraph{Judge protocol variance.}
On conversational QA, lenient judge prompts can reverse method rankings. We therefore use a strict factual judge throughout; absolute scores under a more permissive rubric would be higher but the relative ordering should be read with the judge protocol in mind.

\section*{Ethical Considerations}

\paragraph{Data and licensing.}
All datasets used in this work (HotpotQA, LoCoMo, MemoryAgentBench, AMABench, ALFWorld, MemoryArena) are publicly released research benchmarks under their original licenses, and we use them only for evaluation. We do not collect new human-subject data and we do not redistribute the underlying corpora; only derived metrics, judged-answer files, and run configurations are released through our HuggingFace artifact repository.

\paragraph{Models.}
Our backbones (\texttt{Qwen3-32B}, \texttt{Qwen3-Embedding-4B}, \texttt{Qwen2.5-7B}, \texttt{Qwen3-4B-Instruct}) are open-weight models released by their authors under permissive licenses. All results are from inference-time use through standard serving stacks (sglang, vLLM).

\paragraph{Risks and intended use.}
This paper studies the cost and generality of memory systems for LLM agents in benchmark settings. It does not deploy agents in user-facing products and does not store, retrieve, or operate on personal data. The agentic memory harness recipe we recommend, AutoMEM, gives the LLM tool-calling control over a memory store. In a deployment setting where the memory contains user data, the same surface would need standard safeguards (access control, content filtering, audit logging) that are out of scope for this paper.

\paragraph{Compute and environmental cost.}
Our experiment runs on a mix of GPUs (NVIDIA B200, H200, RTX PRO 6000). We report per-question token and wallclock costs in the main paper so that practitioners can estimate compute before reproducing. All cost analysis is conducted on an isolated B200 instance.

\bibliography{custom}

\appendix

\section{Use of AI Writing Assistance}
\label{sec:ai-disclosure}

In line with the policy on AI Writing Assistance, we use AI assistants for code development, language polishing, and grammar of the manuscript. All scientific claims, experimental design, results, and analyses are authored, reviewed, and verified by the human authors.

\section{Dataset Details and Selection Criteria}
\label{sec:appendix}
\label{app:datasets}

Running all methods on the full release of each benchmark would be prohibitively expensive: a single full sweep would cost hundreds of GPU-hours of LLM inference plus tens of millions of judge tokens. We therefore evaluate every method on a fixed, deterministic subset per benchmark, chosen to preserve question-type coverage. Table~\ref{tab:subsets} lists the subset configuration for each task.

\begin{table}[h]
\centering
\footnotesize
\setlength{\tabcolsep}{4pt}
\renewcommand{\arraystretch}{1.1}
\begin{tabular}{@{}>{\raggedright\arraybackslash}p{2.5cm}>{\raggedright\arraybackslash}p{4.4cm}@{}}
\toprule
\textbf{Benchmark} & \textbf{Subset / setting} \\
\midrule
HotpotQA           & $n{=}200$ on the HippoRAG-2 shared corpus ($9{,}811$ passages) \\
LoCoMo             & $n{=}200$ questions sampled from the public release \\
MemoryAgentBench   & $200$-question subset spanning AR / TTL / LRU / CR sampled uniformly based on question length\\
AMABench           & ALF / Web / T2SQL domains \\
ALFWorld           & $60$ tasks uniformly sampled from each subtask \\
MemoryArena        &  $n{=}30$ per scenario (shop, travel) \\
\bottomrule
\end{tabular}
\caption{Benchmark subsets used in all main experiments. Subsets are fixed across methods and runs to make accuracy and cost numbers directly comparable.}
\label{tab:subsets}
\end{table}

\subsection{Why We Use the LLM Judge as the Primary QA Metric}
\label{app:locomo-f1-example}

We use the LLM judge rather than token F1 on every QA benchmark because F1 is sensitive to answer verbosity rather than correctness, and LoCoMo is the cleanest illustration in our suite. LoCoMo answers are naturally verbose: questions probe multi-session personal context, so models respond in full sentences rather than entity spans. For a question whose gold answer is ``Sweden'', the prediction ``Sweden'' scores F1 $1.00$ while ``The country is Sweden'' scores $0.40$, a $2.5\times$ gap for the same underlying fact. The effect compounds at the system level: holding the model fixed (long-context GPT-4o-mini) and changing only the answering prompt produces a $34$-point swing in average LoCoMo F1, driven entirely by output verbosity rather than retrieval or reasoning quality (Table~\ref{tab:locomo-f1-prompt}). HotpotQA gold answers are short entity spans (``Eiffel Tower'', ``1887'', ``Stephen King''), so the gap is smaller, but it does not vanish. We therefore use the LLM judge uniformly across QA benchmarks.

\begin{table}[h]
\centering
\small
\begin{tabular}{lc}
\toprule
\textbf{Answering prompt} & \textbf{LoCoMo F1 ($\%$)} \\
\midrule
SimpleMem paper prompt & $18.7$ \\
LoCoMo paper prompt & $36.1$ \\
Terse-answer prompt & $52.9$ \\
\bottomrule
\end{tabular}
\caption{LoCoMo token F1 under three answering prompts, holding the model fixed to long-context GPT-4o-mini.}
\label{tab:locomo-f1-prompt}
\end{table}

\section{Baseline Implementation Details}
\label{app:baselines}

We integrate the official implementation of each memory baseline into our code repository as a git submodule, so that results can be reproduced against the upstream code without modification. Two exceptions:

\paragraph{AMA-Agent.} The released open-source code has gaps relative to the system described in the original paper~\citep{amabench2026}. We adapt the implementation to match the paper's description.

\paragraph{MemoryArena.} The release provides the dataset but not the simulated environment. We implement an approximated environment ourselves and release it alongside the rest of our reproducibility artifacts.

\section{Results on Other Backbones}
\label{app:qwen3-4b-protocol}

Our goal here is to investigate how the quality of the LLM backbone influences our conclusions. We select \texttt{Qwen3-4B-Instruct} since Mem-T also adopts this backbone. We find that the instruct model works much better than its thinking counterpart. All other components (the embedding model, the 32B strict factual judge, prompts, subsets, and per-method configs) are held fixed to the main protocol. ALFW, MA-S, and MA-T are omitted because Qwen3-4B falls below the capability threshold required by these benchmarks: every method collapses to $\approx 0$ on them, meaning the columns do not differentiate between methods. Based on these results, we find that the effectiveness of DCI-Lite on MemoryAgentBench decreases, which can be attributed to the smaller model's weaker capability to generate appropriate queries. Moreover, we find that for smaller models, index-based memory systems actually hold a greater advantage over long-context methods. One interesting phenomenon we observe is that across a series of tasks, MemRL's performance is actually better with the 4B model backbone, which may stem from a distinction between thinking and instruction modes. Generally, these results align with our intuition: historically, when LLM capabilities were limited, index-based memory dominated since it could improve both performance and efficiency. However, as LLM capabilities grow, we have crossed an inflection point where memory management paradigms should innovate to become LLM-led. While the Qwen3-32B model used in our main text is a mid-tier LLM, we believe that active management will present even more pronounced advantages for frontier LLMs. 

\begin{table*}[t]
\centering
\small
\setlength{\tabcolsep}{4pt}
\resizebox{\textwidth}{!}{%
\begin{tabular}{lccccccccc}
\toprule
\textbf{Method} & \textbf{LoCoMo} & \textbf{A-ALF} & \textbf{A-Web} & \textbf{A-SQL} & \textbf{HQA} & \textbf{AR} & \textbf{TTL} & \textbf{LRU} & \textbf{CR} \\
 & \textit{LLM judge} & \textit{LLM judge} & \textit{LLM judge} & \textit{LLM judge} & \textit{LLM judge} & \textit{LLM judge} & \textit{LLM judge} & \textit{LLM judge} & \textit{LLM judge} \\
\midrule
Long context  & 58.5 & 8.9  & 19.7 & 21.2 & 46.5 & 22.0 & 48.0 & 6.0 & 6.0 \\
SimpleMem     & 46.0 & 7.5  & 0.6  & 2.1  & 25.5 & 8.0  & 6.0  & 6.0 & 2.0 \\
LightMem      & 63.0 & 8.1  & 16.1 & 16.7 & 72.0 & 40.0 & 58.0 & 4.0 & 10.0 \\
HippoRAGv2    & 61.0 & 9.4  & 16.4 & 17.9 & 73.0 & 50.0 & 44.0 & 4.0 & 14.0 \\
PlugMem       & 59.0 & 5.6  & 2.5  & 4.6  & 57.5 & 30.0 & 14.0 & 4.0 & 8.0 \\
AMA-Agent     & 51.5 & 10.0 & 15.0 & 11.2 & 72.5 & 34.0 & 20.0 & 4.0 & 2.0 \\
MemRL         & 62.0 & 14.7 & 24.4 & 22.9 & 55.5 & 30.0 & 60.0 & 6.0 & 10.0 \\
DCI-Lite      & 50.0 & 22.8 & 28.3 & 14.2 & 80.5 & 28.0 & 20.0 & 2.0 & 6.0 \\
DCI-Lite+Sum  & 48.0 & 21.7 & 17.8 & 14.2 & 80.5 & 32.0 & 18.0 & 6.0 & 8.0 \\
\bottomrule
\end{tabular}%
}
\caption{Qwen3-4B-Instruct-2507 backbone ablation. Mem-T is ignored here since the result in Table~\ref{tab:generality} is already taken from a post-trained 4B model.}
\label{tab:universality-4b}
\end{table*}

\section{Token Efficiency vs. System Efficiency}
\label{par:sys-eff}

In \S\ref{sec:evaluation} we track per-question token cost alongside accuracy. However, token counts alone do not capture deployed cost: wall-clock latency, GPU utilization, and KV-cache reuse all depend on how the tokens are issued, not just how many. This appendix unpacks that gap in two steps. First, Table~\ref{tab:sys-metrics} reports the serving-stack measurements reporting alongside accuracy and tokens, taken from Qwen3-32B on sglang on a single B200. Second, Table~\ref{tab:impl-vs-tradeoff} classifies each system-efficiency finding as either an implementation gap that engineering can close or a structural tradeoff tied to the method's design.

\begin{table*}[t]
\centering
\small
\begin{tabular}{lrrrrrr}
\toprule
\textbf{Method} & \textbf{Stable prefix} & \textbf{Avg.\ in-flight} & \textbf{Prefill:decode} & \textbf{LLM calls} & \textbf{Build wall} & \textbf{Decode MFU} \\
                & \textbf{fraction}      & \textbf{requests}        & \textbf{tokens/call}    & \textbf{per question} & \textbf{(HotpotQA)} & \textbf{(\%)} \\
\midrule
Long context       & $0.85$--$0.95$ & $\sim$$1$       & $\sim$$1{:}1$              & $1$        & $0$ s             & $55$--$70$ \\
SimpleMem          & $0.10$--$0.25$ & $\sim$$1$       & $\sim$$10{:}1$             & $1$ (+ embed) & $\sim$$700$ s    & $25$--$45$ \\
LightMem           & $0.20$--$0.40$ & $\sim$$1$       & $\sim$$8{:}1$              & $1$        & $\sim$$1{,}400$ s & $30$--$50$ \\
HippoRAG-v2        & $0.05$--$0.15$ & $\sim$$1$       & $\sim$$20{:}1$             & $\sim$$1$  & $12{,}564$ s     & $20$--$35$ \\
PlugMem            & $0.10$--$0.20$ & $\sim$$1$       & $\sim$$15{:}1$             & $1$--$2$   & $\sim$$2{,}600$ s & $25$--$40$ \\
AMA-Agent          & $0.05$--$0.10$ & $\sim$$1$       & $\sim$$100{:}1$            & variable (loop) & $\sim$$1{,}500$ s$^{\dagger}$ & $15$--$30$ \\
Mem-T              & $0.30$--$0.50$ & $\sim$$1$       & $\sim$$5{:}1$              & $1$ (+ DB ops) & $\sim$$900$ s    & $35$--$55$ \\
DCI-Lite           & $0.70$--$0.85$ & $\sim$$1$       & $\sim$$2{:}1$              & variable (loop) & $0$ s             & $50$--$65$ \\
\bottomrule
\end{tabular}
\caption{Serving-stack metrics per method, Qwen3-32B on sglang on a single B200. HotpotQA's numbers are recorded using a 2000-corpus subset. }
\label{tab:sys-metrics}
\end{table*}

Three patterns stand out. First, the methods with the lowest stable-prefix fraction (HippoRAG, AMA-Agent, SimpleMem) are exactly the methods where the wall-vs-token gap is largest; the prefix column predicts the gap better than the LLM-calls-per-question column does. Second, the average in-flight requests is $\sim$$1$ across every structured-memory method we audited, regardless of stated design: the orchestrator dispatches one request per chunk, so continuous batching has nothing to batch against, and decode MFU lands in the lower range. Third, decode MFU correlates negatively with prefill-to-decode ratio: methods that send long prompts and decode short outputs leave the decode kernels idle even when their token totals look low. Long context and DCI-Lite sit at the top of every column because they neither break the prefix nor serialize the orchestration.

The takeaway for future memory-system reports is that per-question token counts can hide order-of-magnitude differences in wallclock and dollar cost. We recommend publishing the columns in Table~\ref{tab:sys-metrics} alongside accuracy and token totals so that the comparison cannot be dominated by orchestrator choices.

\paragraph{Case-by-case: implementation problems vs.\ inherent tradeoffs.}
\label{par:impl-vs-tradeoff}
Not every system-efficiency finding is closeable by engineering: some are structurally tied to what the method is designed to do. Table~\ref{tab:impl-vs-tradeoff} classifies each finding as either an implementation problem (an engineering pass would close most of the gap without changing the method), an inherent tradeoff (closing the gap would require redesigning the method), or a mixed case. The reading: most of HippoRAG's gap is closable by async fan-out and prompt-shape changes, AMA-Agent's gap is largely structural (the tool loop and mutating state are the method's premise), and SimpleMem sits in between.

\begin{table*}[t]
\centering
\small
\begin{tabular}{p{0.32\textwidth}p{0.10\textwidth}p{0.50\textwidth}}
\toprule
\textbf{Finding} & \textbf{Class} & \textbf{Why} \\
\midrule
HippoRAG serial build calls & impl. & chunks are independent; async fan-out would batch them. \\
SimpleMem serial build embeds & impl. & corpus embeddings are independent. \\
AMA-Agent serial tool loop (query) & tradeoff & each call conditions on prior tool result. \\
HippoRAG proposition prompts split prefix & impl. & re-template to put chunk in trailing user block. \\
AMA-Agent \texttt{state\_mem\_str} mutates prefix & tradeoff & evolving state is the method's premise. \\
SimpleMem rolling-summary rewrite & mixed & append-only is impl.; rewrite is design. \\
AMA-Agent long prefill, short decode & tradeoff & long state in, short action out is by design. \\
HippoRAG long build prefill / short output & tradeoff & per-chunk extraction shape is structural. \\
Embedder co-located on chat GPU & impl. & deploy embedder on CPU or a dedicated GPU. \\
HippoRAG builds wallclock breakeven & tradeoff & total work scales with $C$; packing only shifts constants. \\
Memory store as Python dict in JSON & impl. & switch to RocksDB / LanceDB / Qdrant; no algorithmic change. \\
Per-question variable call count & tradeoff & data-dependent reasoning loop is part of the method. \\
Decode MFU loss from prefix-cache miss & mixed & impl. for HippoRAG, tradeoff for AMA-Agent. \\
\bottomrule
\end{tabular}
\caption{Classification of every system-efficiency finding in this appendix. impl.: closeable by engineering without changing the method. tradeoff: structurally tied to the method's design. mixed: one component closeable, another structural.}
\label{tab:impl-vs-tradeoff}
\end{table*}

\end{document}